\documentclass[runningheads]{llncs}
\usepackage[T1]{fontenc}
\usepackage{graphicx}
\usepackage{float} 
\usepackage{xcolor}
\usepackage{comment}
\usepackage{multirow}
\usepackage[english]{babel}

\usepackage{wrapfig}

\usepackage[skip=10pt plus1pt]{parskip}

\usepackage{cite}

\begin{document}
\title{Assessing the Impact of Noise on Quantum Neural Networks: An Experimental Analysis}
\titlerunning{Impact of Noise on Quantum Neural Networks}
%
\author{Erik Terres Escudero\inst{1}\orcidID{0009-0003-9781-7657}  \and Danel Arias Alamo\inst{1}\orcidID{0000-0002-6586-346X} \and
Oier Mentxaka Gómez\inst{1}\orcidID{0009-0004-9667-5834} \and 
Pablo García Bringas\inst{1}\orcidID{0000-0003-3594-9534}}
\authorrunning{Terres et al.}
%
\institute{Deusto University, Bilbao, Spain\\ \email{\{danel.arias, oiermentxaka\}@opendeusto.es} \\ \email{\{e.terres, pablo.garcia.bringas\}@deusto.es}}

%
%

%
\maketitle              
\begin{abstract}
In the race towards quantum computing, the potential benefits of quantum neural networks (QNNs) have become increasingly apparent. However, Noisy Intermediate-Scale Quantum (NISQ) processors are prone to errors, which poses a significant challenge for the execution of complex algorithms or quantum machine learning. To ensure the quality and security of QNNs, it is crucial to explore the impact of noise on their performance. This paper provides a comprehensive analysis of the impact of noise on QNNs, examining the Mottonen state preparation algorithm under various noise models and studying the degradation of quantum states as they pass through multiple layers of QNNs. Additionally, the paper evaluates the effect of noise on the performance of pre-trained QNNs and highlights the challenges posed by noise models in quantum computing. The findings of this study have significant implications for the development of quantum software, emphasizing the importance of prioritizing stability and noise-correction measures when developing QNNs to ensure reliable and trustworthy results. This paper contributes to the growing body of literature on quantum computing and quantum machine learning, providing new insights into the impact of noise on QNNs and paving the way towards the development of more robust and efficient quantum algorithms.

\end{abstract}

\section{Introduction}

In recent years, quantum computing has made remarkable progress, and its potential advantages over classical computing have become increasingly apparent. Quantum neural networks (QNNs) are a promising approach to quantum artificial intelligence that leverage the unique properties of quantum systems to achieve exponential memory capacity, scalability, and faster learning. Several researchers have proposed QNNs as a possible alternative to classical neural networks, highlighting their potential benefits \cite{Ezhov2000, GUPTA2001355, Schuld2014}.

Noisy Intermediate-Scale Quantum (NISQ) processors have made quantum systems with hundreds of qubits available, which is a significant milestone for quantum computing. However, the results generated by these systems are still noisy and prone to errors, which poses a challenge for the execution of complex algorithms or quantum machine learning. The combination of the inherent instability of neural networks with the inconsistency and error-proneness of quantum computing creates a challenging landscape for researchers to navigate. Nevertheless, these challenges present a unique opportunity for researchers to explore new methods and techniques to address the limitations of both quantum computing and neural networks.

Ensuring the quality and security of quantum neural networks is a crucial step in guaranteeing that production-ready industry models perform as intended, requiring high accuracy and robustness against noisy data. Potential quantum errors could be exploited by malicious agents to manipulate the output of the network, leading to inaccurate predictions or faulty decisions. To safeguard against such attacks, quantum software development must adopt a rigorous approach with strict quality criteria and error-free execution \cite{arias_lets_2023}.


Our work provides a comprehensive analysis of the impact of noise on quantum neural networks. We examine the Mottonen state preparation algorithm \cite{mottonen_transformation_2004} under various noise models and study the degradation of quantum states as they pass through multiple layers of quantum neural networks. Additionally, we evaluate the effect of noise on the performance of quantum neural networks and highlight the challenges posed by noise models in quantum computing.

The structure of this paper is organized as follows. In Section 2, we review the existing literature and highlight the key contributions of prior research in this area. In Section 3, we describe our experimental approach and methodology for analyzing the effects of noise on quantum neural networks. In Section 4, we present the empirical findings of our analysis. In Section 5, we discuss the implications of our findings and their significance. Finally, in Section 6, we draw conclusions and suggest future research directions in this field.






\subsection{Quantum Neural Networks}

Quantum computing leverage qubits, which grants it with unique properties such as superposition and entanglement. In order to operate, quantum computers make use of quantum gates (e.g. rotation $R_x$, and CNOT/$C_x$). Even if these properties make Quantum computing powerful, the current state of the art in quantum computers are NISQ (Noisy Intermediate-Scale Quantum) which suffer from various types of noise and errors that make them less reliable if not correctly used. To mitigate this, researchers are working on different physical improvements or algorithms, such as Quantum error correction algorithms.

Quantum neural networks are a special type of neural network which leverage the power of these quantum properties to learn complex data models and solve problems. To implement such networks, Variational Quantum Circuits (VQC) are constructed by a series of gates with trainable parameters which can be tuned. 

These circuits approximate classical learning by emulating the internal structure of classical neural networks using a construction of CNOT and Rotation Gates. This layer structure, known as a strongly entangled layer, is similar to a classical layer. CNOT connections represent synapse connections, while rotations on the layer represent weighted sum transformations.\cite{Schuld_2020}

While more complex network structures, such as quantum activation functions or quantum recurrent networks, have been proposed, their high implementation complexity makes them impractical for this work. Therefore, we will rely on standard rotation/entangled layered networks \cite{henderson_quanvolutional_2019, maronese_quantum_2022, bausch_recurrent_2020}.

\subsection{The Challenges on Measuring Error on Quantum Neural Networks }

The challenges surrounding quantum neural networks are multifaceted, stemming from both the early state of quantum computing and the complexity of neural networks. One key challenge is the inherent error proneness of quantum hardware, which limits the viability of deep QNNs.  Due to the current noise in quantum computers, the circuits on the hardware can only have limited depth, restricting the size and complexity of QNNs that can be developed. Developing deeper QNNs demands multiple layers of quantum gates, which increases the impact of errors.

Another significant challenge is the lack of a clear theoretical framework for QNNs. This makes it difficult to understand and quantify the errors in these systems and develop effective error correction techniques. Furthermore, the development of such techniques is also challenging due to the complex interplay between the quantum hardware and the neural network algorithms. Therefore, addressing these challenges is necessary to enable the development of robust QNNs that can effectively solve complex problems.

\section{Related Work}

Quantum Variational Circuits are a type of parametrized quantum circuit that use a hybrid learning methodology for training quantum neural networks \cite{cerezo_variational_2021}. Quantum data is processed by the circuit, while the output and the training is done by classical training optimization techniques, such as backpropagation. This approach makes them a powerful tool for solving a wide range of problems in fields such as supervised classification \cite{henderson_quanvolutional_2019, rebentrost_quantum_2014, Hur2022} and reinforcement learning \cite{lockwood_reinforcement_2020, lockwood_playing_nodate}.

Two primary techniques are commonly utilized for initializing data into the circuit, namely Angle Embedding and Amplitude Embedding. Whilte angle-based states make use of fewer gates, their information storage capacity scales linearly with the number of qubits, which makes them unsuitable for handling high-dimensional data. On the other hand, amplitude embedding techniques, such as Mottonen state preparation algorithm, enable exponentially greater data dimensionality at the expense of an exponentially larger number of required gates \cite{mottonen_transformation_2004}.

Despite the potential benefits of quantum neural networks, the presence of noise in NISQ computers can reduce their learning capacity by causing barren plateaus, which result in a vanishing gradient and limit the learning capabilities of these systems \cite{wang_noise-induced_2021}. Although several Quantum Error Correction techniques exist, they do not guarantee error-free execution of quantum circuits \cite{roffe_quantum_2019}. However, recent research suggests that the presence of some low level of noise may help avoid saddle points and ensure the model's convergence \cite{liu_noise_2022}. In order to achieve quantum advantage, it is essential to ensure that quantum computers are robust against environmental noise and gate errors \cite{huang_quantum_2022}.

\section{Methodology}

The present study aims to tackle three fundamental challenges in QNNs: (1) how environmental noise and gate error affects the state of a quantum system as it passes through a quantum neural network, (2) how resilient amplitude state preparation algorithms are to noise, and (3) how noise impacts the performance of pre-trained quantum neural networks.


To evaluate the impact of noise on the quantum state under increasing layers, we will prepare uniformly initialized quantum neural networks and run several executions with random weights to evaluate the degradation of the state. We will analyze the rate of degradation with respect to two baselines: the resultant state of a noise-free evaluation on the same circuit and the expected convergence state of the system under high noise.

Regarding the second problem, we will evaluate the resilience of amplitude state preparation algorithms to noise by analyzing the effect of different noise models on the prepared state. We will provide visual information of the resultant state and a later comparison of the effect under quantum neural networks.



For the third problem, we will first train multiple quantum neural networks in a noise-free environment and then evaluate their performance under various noisy models provided by IBM Quantum. We will use the MNIST dataset as a benchmark and measure the degradation in performance caused by the noise. To better understand the impact of noise on classification performance, we will conduct experiments with different class splits and analyze how the space of the classification is affected by the noise perturbation.

To avoid any bias in the results, we will use multiple noise models with different specifications to evaluate the impact of noise on QNNs. This will allow us to examine how different noise models affect QNNs in unique ways, ensuring that the results are not influenced by a single noise model.

Overall, our approach involves training and testing quantum neural networks under different noisy conditions, using appropriate metrics to evaluate performance, and comparing the results to identify the impact of noise on the quantum neural network.


\subsection{Experimental Setup}

To provide accurate results on the quantum simulations, real quantum machine specifications will be used. Specifically, we will make use of the AER simulator, which mimics the execution of the quantum circuits on actual devices, providing reliable and precise results. In order to minimize bias and ensure the quality of the work, we have selected four distinct quantum systems to extract their specifications for the simulator: IBM Cairo, IBMQ Guadalupe, IBM Hanoi, and IBMQ Mumbai. These simulators were chosen based on their compatibility with our research requirements, ensuring a minimum of 8 qubits and the capability to sample from a variety of quantum models. 

We chose the MNIST dataset to test the impact of noise on trained quantum machine learning models' inference capacity. We will test all models for 2 (0-1), 4 (0-3), and 10 (0-9) classes to investigate whether the number of classes affects the error rate. Given that the input dimension is 784, we will use amplitude encoding because angle embeddings are not feasible. To reduce redundancy and address memory restrictions, we will reduce the data to 14x14 (196) dimensions through max pooling (2x2 kernels) with strides (2x2). We will then project the 196 dimensions to the 256 states of an 8 qubits system, setting the extra states to zero.

We will use Pennylane as the main quantum machine learning library and Qiskit as a backend for quantum circuit simulation. The circuits will have 8 qubits, and the networks will follow a standard structure. We will prepare the initial state with a Mottonen state preparation circuit followed by a sequential chain of strongly entangled layers. In total, we will prepare 5 different networks, with 1, 3, 5, 7 and 9 layers respectively. Measurements will be given as the average state of each qubit at the end of the circuit. To account for a variable number of classes and since the quantum circuits contain 8 qubits, we will connect the output of the quantum network to a classical dense classification layer.

To train the quantum neural networks, we will utilize the Pennylane lightning plugin with Tensorflow as the interface, following a supervised learning approach. The 5 networks will be trained on the MNIST dataset, split into three categories: 0-1, 0-3, and 0-9, for 1, 2, and 4 epochs, respectively. We will use the Adam optimizer with a learning rate of 0.01 and a categorical cross-entropy loss function. The adjoint optimization algorithm will be employed as the backpropagation algorithm, as it is both fast and reliable. The training will use 600 shots on the quantum circuit and a batch size of 16 to reduce the statistical noise in the measurement outcomes.


\section{Results}

\begin{wrapfigure}[13]{r}{0.5\textwidth}
  \centering
  \includegraphics[width=0.48\textwidth]{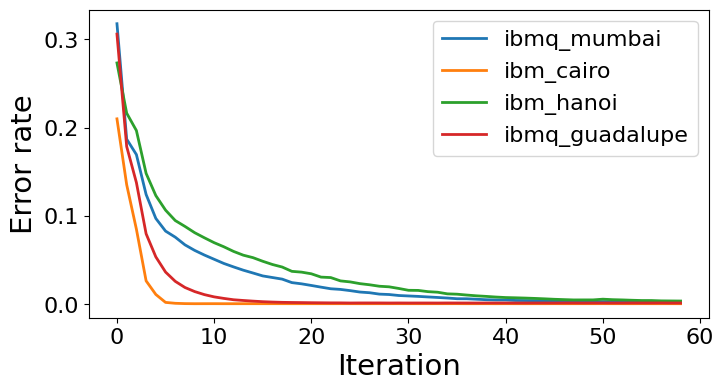}
  
  \caption{$\chi^2$ distance with respect to a uniform distribution per iteration, up to 60 iterations for the 4 specified backends. }
  \label{fig:plot-noise}
\end{wrapfigure}

The experimental result reveals that noise in IBM quantum systems causes latent states to converge towards a uniform distribution, rendering the system unable to distinguish between real states and the uniform states. As depicted in Figure \ref{fig:plot-noise}, the degradation rate of the system follows an exponential decay. The rate of degradation of the state strongly varies with the chosen noise model chosen, with IBM Hanoi and IBMQ Mumbai allowing for deeper networks without impactful degradation, taking up to 50 steps to fully converge towards a uniform distribution, while IBM Guadalupe takes up to 10 layers and IBM Cairo takes up to 5 layers. 

\begin{figure}[hh]
    \centering
    \includegraphics[width=\textwidth]{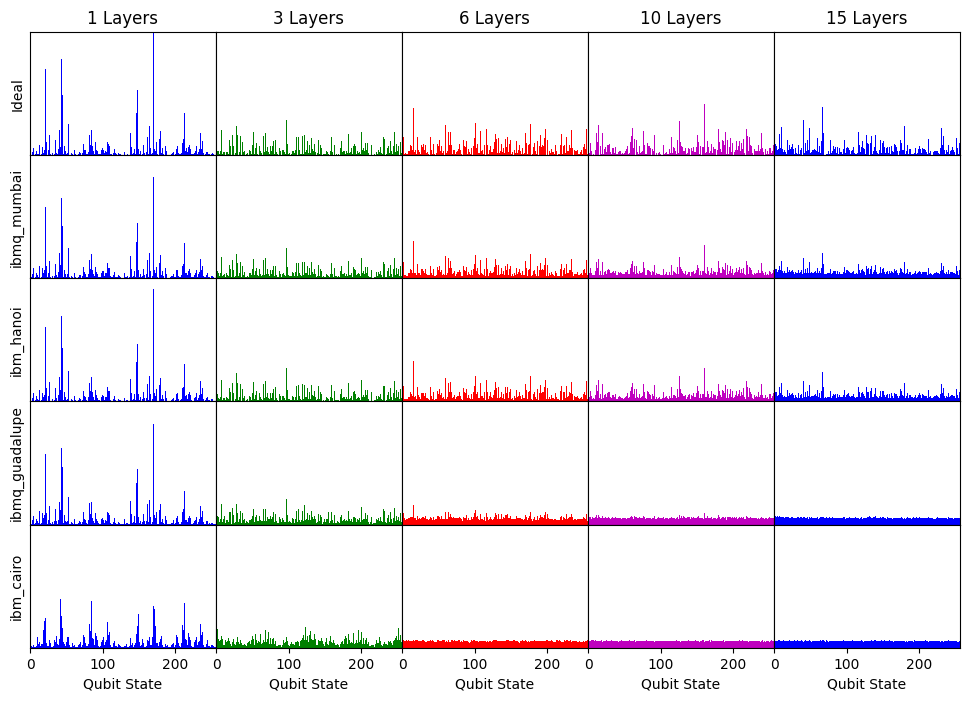}
    \caption{Evolution of the qubit state under the 4 specified noise models in several quantum neural networks with different amount of random weighted layers (1, 3, 6, 10 and 15 layers respectively) .}
    \label{fig:noisy-state}
\end{figure}

Although intrinsic noise perturbs the state of quantum systems, the overall distribution of the data appears to remain. As shown in Figure \ref{fig:noisy-state}, for example, on IBM Hanoi or IBMQ Mumbai at layer 15, while a clear uniform floor has been formed, the highest states of the distribution still retain their order and relative magnitude with the original state. However, as the depth increases, the magnitude of conservation of the real state decreases until the distribution is equal to the uniform distribution.

It is worth noting that IBM noise models are updated regularly, and their specifications are updated online daily. During the development of the paper, the noise models cycled between lower and higher noise models, with some unable to hold the distribution under even one layer. Therefore, it is essential to ensure that the model's specifications meet certain robustness criteria before using noisy models in production, especially as no alert is triggered when the noise models are updated.



\begin{figure}[hh]
    \centering
    \includegraphics[width=0.75\textwidth]{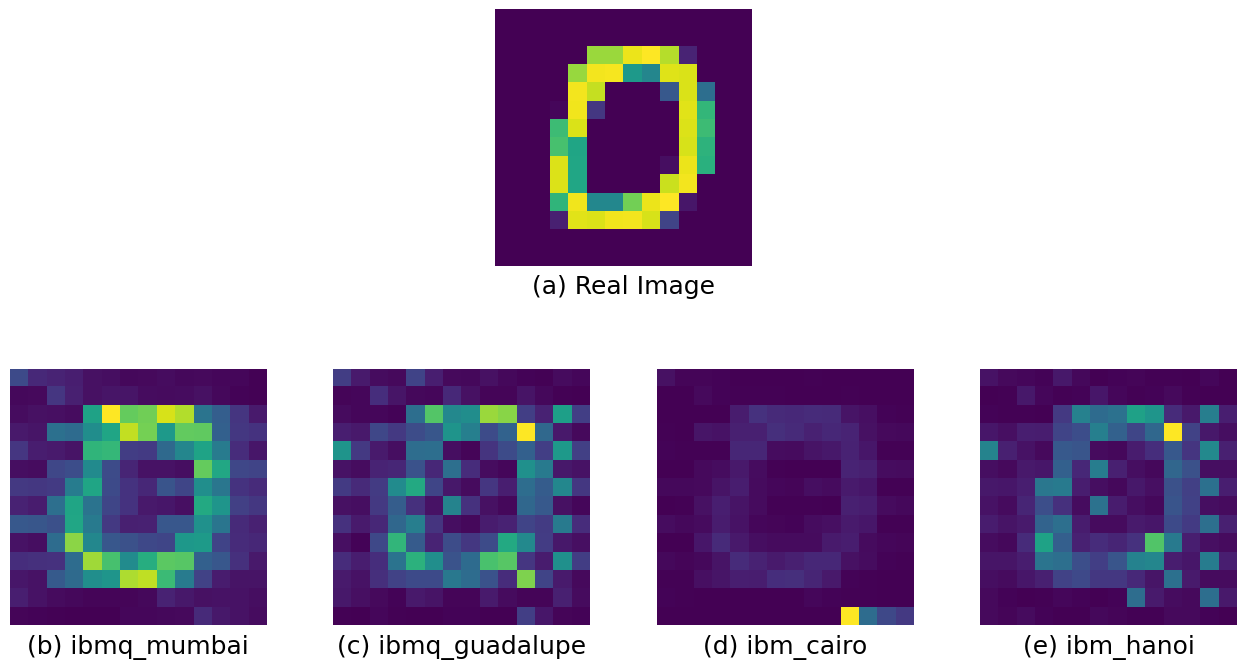}
    \caption{Mottonen State Preparation on the specified noise models from different quantum computers: (a) Real image, (b) IBMQ Mumbai noise model effect, (c) IBMQ Guadalupe noise model effect, (d) IBM Cairo noise model effect, (e) IBM Hanoi noise model effect.}
    \label{fig:noisy-images}
\end{figure}

Our analysis of Amplitude Embedding algorithms revealed that high noise errors in gates or readout resulted in a faulty distribution of the state, causing specific pixels of the image to have sharper values than their neighbors. Figure \ref{fig:noisy-images} provides a clear example of this behavior in IBM Cairo, where a faulty CNOT with an error rate of 1, acting between qubits 0 and 1, creates a sharp pixel in the background. This pixel absorbs half of the distribution of the state, maintaining the shape of the zero in the background but completely altering the distribution of the data.

In contrast, IBM Hanoi, IBMQ Guadalupe, and IBMQ Mumbai were able to prepare the state in a way that was still visible. Although IBMQ Guadalupe added a higher degree of background noise, the most important pixels were still present in the image. Among the three, IBMQ Mumbai was the most precise noise model in preparing quantum states by providing an evenly distribute state through the expected pixels while keeping a moderate background noise. Yet, as it can be seen in Figure \ref{fig:plot-noise}, the background noise in IBMQ Mumbai is stronger than IBM Hanoi's, degrading the state of the circuit faster. IBM Hanoi, while not having the best distribution over the pixels, contains the most robust noise distribution over the different backends.

As the data is encoded through binary CNOT gates, most of the noise in the images can be clearly attributed to binary location. This trend is visible in the results obtained from IBMQ Guadalupe and IBM Hanoi, where a trace of high intensity pixels can be seen on the even pixels on the right side of the images. It is important to note that this noise distribution behaves differently from classical noise, which is uniformly distributed throughout the image. The noise in quantum data follows a clear trend to focus on states which are divisible by different powers of two. This characteristic of quantum noise should be taken into account when dealing with data preparation or noise correction in future algorithms.

The results presented in Table \ref{tab:results-noisy-model} clearly demonstrate the impact of noise levels on model accuracy. In particular, the noise levels in IBM Cairo are significant enough to severely limit the model's learning ability, as evidenced by the deformed pixel shown in the state preparation process. This shift in the data leads to a significant decrease in accuracy, as expected.

While IBMQ Guadalupe is a less noisy model compared to IBM Cairo, it still struggles to maintain accuracy beyond a one-layer neural network and quickly degrades towards a random model. On the other hand, IBM Hanoi and IBMQ Mumbai, which are the least noisy models, are able to maintain performance over different numbers of layers, but still suffer a noticeable accuracy loss.

Our analysis showed that certain noise models had a greater impact on QNNs trained with different numbers of layers. This can be attributed to the fact that noise models affect specific gates, readouts, or connections with varying degrees of strength. As a result, different weight sets trained on the same data may be impacted differently by the same noise model, resulting in varying levels of performance degradation.  

Additionally, we observed a significant decrease in accuracy when increasing the number of classes on the network. For instance, IBMQ Mumbai, which was able to accurately solve the 2 and 4 classes split, struggled when dealing with 10 classes, failing to reach 50\% accuracy in any number of layers. Similarly, IBM Hanoi, which performed better initially, also suffered significantly, with only one model achieving 70\% accuracy.

\begin{table}
    \centering
    \caption{Accuracy of the pre-trained QNNs for the specified noise models and number of layers.}
    \begin{tabular}{|l|p{1.5cm}|p{1.5cm}|p{1.5cm}|p{1.5cm}|p{1.5cm}|}
        \hline
        Noise Model &  1 Layer & 3 Layers & 5 Layers & 7 Layers & 9 Layers \\ \hline
        \multicolumn{6}{|c|}{Classes 0-1} \\
        \hline
        IBM\_Cairo &  38.16\% & 36.53\% &  38.05\% &  55.56\% & 55.63\% \\
        IBMQ\_Guadalupe & 47.79\% & 50.31\% & 38.05\% &  55.71\% & 61.95\%\\
        IBMQ\_Mumbai & 94.69\% &  97.34\% & 99.12\% & 98.78\% & 95.58\% \\
        IBM\_Hanoi & 99.10\% & 99.43\% & 99.27\%  & 99.12\%   & 99.89\%  \\
        Base & 96.02\% & 98.73\% & 99.31\%& 99.33\% & 99.29\% \\
        \hline
        \multicolumn{6}{|c|}{Classes 0-3} \\
        \hline
        IBM\_Cairo &  26.43\% & 19.10\% &  26.5\% & 30.23\% & 26.49\%\\
        IBMQ\_Guadalupe & 39.13\% & 24.56\% &  23.64\% &  25.45\% & 28.12\%\\
        IBMQ\_Mumbai & 80.67\% &  55.47\% & 89.71\% & 89.53\% & 78.29\%\\
        IBM\_Hanoi & 88.95\% & 89.67\% & 89.09\%  &  90.18\% & 90.38\% \\
        Base & 84.23\% & 92.79\%& 93.86\%& 94.56\% & 94.74\%\\
        \hline
        \multicolumn{6}{|c|}{All Clases} \\
        \hline
        IBM\_Cairo &  9.83\% & 10.22\% &  10.57\% &  9.70\% & 11.62\% \\
        IBMQ\_Guadalupe & 25.74\% & 17.04\% &  10.57\% &  21.08\% & 13.12 \%\\
        IBMQ\_Mumbai & 44.87\% &  29.03\% & 17.02\% & 37.56\% & 29.52\%  \\
        IBM\_Hanoi & 54.22\% & 69.51\% &  70.46\% &  68.16\% &  52.47 \% \\
        Base & 59.46\% & 70.45\% & 72.74\% & 78.22\%& 79.43\% \\
        \hline
    \end{tabular}
    \label{tab:results-noisy-model}
\end{table}
\vspace{-0.3cm}
\section{Discussion}

In this study, we aimed to investigate the impact of noise on quantum neural networks in IBM quantum systems. Our findings suggest that the presence of noise in quantum systems causes latent states to converge towards a uniform distribution, making it difficult to distinguish between real states and uniform states. The rate of degradation of the state strongly depends on the chosen noise model, with IBM Hanoi and IBMQ Mumbai being the least noisy models, while IBMQ Guadalupe and IBM Cairo experience a more significant loss of accuracy. However, on initial layers, the distribution of the data appears to retain certain structure, allowing for classical post-processing of the output. Nevertheless, these results highlight the need for noise-robust systems to build deep QNNs reliably.

The analysis on the effect of Amplitude Embedding in different quantum computing environments showed that high noise errors in gates or readout resulted in a faulty distribution of the state, causing specific pixels of the image to have sharper values than their neighbors. This effect can be attributed to the high dependency of the Mottonen State preparation on CNOT gates, with are one of the most error-prone gates. The exponential need of CNOT gates implies a high probability of degradation on noisy quantum systems. Notably, since CNOT gates are binary gates, the error trace observed on the image exhibited a clear binary aspect, where sets of powers of 2 manifested high noise values.

This trend is visible in the results obtained from IBMQ Guadalupe and IBM Hanoi, where a trace of high-intensity pixels can be seen on the even pixels on the right side of the images. This characteristic of quantum noise should be taken into account when dealing with data preparation or noise correction in future algorithms.

The results presented in this study clearly demonstrate the significant impact of noise levels on the accuracy of QNNs. Models with cleaner state preparation achieved better accuracies, and the accuracy of the models was directly related to their ability to retain the distribution of their data from the uniform distribution. These findings highlight the importance of having circuit quality measures in place to assess the stability of QNNs under ongoing noise circumstances. As seen in the table, circuits trained with similar expected accuracy can yield vastly different results when subjected to noise.

The impact of increasing the number of classes on the performance of QNNs is significant. This is due to the nature of QNNs as mathematical functions that map data spaces. As the number of classes increases, the distance between different class spaces decreases, making it easier for any perturbation in the data caused by intrinsic noise to move the latent data from one class to another. Therefore, if the goal is to develop deeper and more complex QNNs, it is crucial to reduce noise to a level where perturbations have even lower thresholds of action. Otherwise, accumulated noise perturbations will inevitably distort the output, leading to incorrect classifications.

Given the high cost of training quantum neural networks on actual quantum computers, the training in this study was conducted on simulators. However, training on a noise-robust quantum computer could reveal valuable insights into the capacity and limitations of QNNs in real-world environments. Therefore, an important future direction would be to extend these results to real quantum computers. Another potential line of research would involve conducting an ablation study on the different noise factors that make up a general noise model in quantum computing, such as T1, T2, and gate errors. Such an analysis could help identify which noise factors are most significant and require the most attention in developing robust QNNs.

\section{Conclusion}


In this investigation, the effect of noise in IBM quantum systems on deep quantum neural networks has been studied. The results indicate that noise in quantum systems causes qubit states to converge exponentially towards a uniform distribution, rendering the system unable to operate with the state. The rate of degradation of the state depends on the chosen noise model, highlighting the need for noise-robust systems to develop deep quantum neural networks reliably. Nonetheless, the fundamental structure of the quantum state remains intact for several layers, indicating the feasibility of developing noise reduction techniques on the quantum output.

The study demonstrated the influence of noise on quantum state preparation, highlighting that noise-tolerant models resulted in improved image representation in the quantum state. Notably, the observed noise in quantum systems differed from classical systems, as it exhibited a pattern aligned with multiples of powers of 2, potentially due to interactions between various CNOT gates and connectivity structures. This unique characteristic of quantum noise should be taken into account in future algorithms for noise correction or data preparation.

The current state of quantum hardware limits the depth of circuits that can be used, making it challenging to build deep QNNs. Different noise models affect QNNs with varying degrees of strength, which can impact their performance differently. Furthermore, increasing the number of classes in a dataset leads to a decrease in accuracy due to the geometrical nature of QNNs as mathematical functions that map data spaces. These findings underscore the importance of developing circuit quality measures to assess the stability of QNNs under noise and the need for future work to explore training on actual quantum hardware.

\section*{Acknowledgements}

The authors would like to acknowledge the partial financial support by Ministry of Science (project QSERV-UD, PID2021-124054OB-C33), and also to the Basque Government (projects TRUSTIND - KK-2020/00054, and REMEDY - KK-2021/00091). Additionally, the authors wish to acknowledge the selfless support from IBM, who generously provided their quantum computing equipment for the project. Finally, it is important to also express gratitude for the support and drive that the regional government of Bizkaia is providing in all matters related to the development of quantum technologies as a driving force for progress of the Society of this historic territory.

\renewcommand{\contentsname}{References}
\renewcommand{\refname}{References}
\bibliographystyle{IEEEtran}
\bibliography{bibliography}

\end{document}